\documentclass[sigconf,10pt,screen]{acmart}

\usepackage{etoolbox}
\usepackage{graphicx}
\usepackage{setspace}
\usepackage{listings}
\usepackage{multicol}
\usepackage{xspace}
\usepackage{enumitem}
\usepackage{booktabs}
\usepackage[labelfont=bf]{caption}
\usepackage{siunitx}
\usepackage{color}
\usepackage{algorithmicx, algorithm}
\usepackage{algpseudocode}
\usepackage{subfigure}
\usepackage{multirow}

\definecolor{Code}{rgb}{0,0,0}
\definecolor{Numbers}{rgb}{0.5,0,0}
\definecolor{Keywords}{rgb}{0,0,0.5}
\definecolor{CrossTrainer}{rgb}{0,0.5,0}
\definecolor{self}{rgb}{0,0,0}
\definecolor{Strings}{rgb}{0,0.63,0}
\definecolor{Comments}{rgb}{0,0.63,1}
\definecolor{Backquotes}{rgb}{0,0,0}
\definecolor{Classname}{rgb}{0,0,0}
\definecolor{FunctionName}{rgb}{0,0,0}
\definecolor{Operators}{rgb}{0,0,0}
\definecolor{Background}{rgb}{0.98,0.98,0.98}
\lstdefinelanguage{Python}{
    numbers=left,
    numbersep=1em,
    xleftmargin=1em,
    framextopmargin=2em,
    framexbottommargin=2em,
    showspaces=false,
    showtabs=false,
    showstringspaces=false,
    tabsize=4,
    breaklines=true,
    basicstyle=\ttfamily\footnotesize\setstretch{1},
    backgroundcolor=\color{Background},
    comment=[l][\color{Comments}\slshape]{\#},
    commentstyle=\color{Comments}\slshape,
    morekeywords={import,from,class,def,for,while,if,is,in,elif,else,not,and,or,print,break,continue,return,True,False,None,access,as,,del,except,exec,finally,global,import,lambda,pass,print,raise,try,assert},
    keywordstyle={\color{Keywords}\bfseries},
    morekeywords=[2]{CrossTrainer},
    keywordstyle=[2]{\color{CrossTrainer}\bfseries},
    literate=*
    {5}{{{\color{Numbers}5}}}1
    {0.01}{{{\color{Numbers}0.01}}}4,
    emph={self},
    emphstyle={\color{self}\slshape},
}

\newcommand*{\reweighting}{loss reweighting\xspace}
\newcommand*{\Reweighting}{Loss reweighting\xspace}
\newcommand*{\sysname}{CrossTrainer\xspace}
\newcommand{\minihead}[1]{{\vspace{.45em}\noindent\textbf{#1.} }}

\newcommand*{\DA}{DA\xspace}

\algnewcommand{\algorithmicor}{\textbf{ or }}
\algnewcommand{\OR}{\algorithmicor}
\def\BibTeX{{\rm B\kern-.05em{\sc i\kern-.025em b}\kern-.08emT\kern-.1667em\lower.7ex\hbox{E}\kern-.125emX}}

\acmPrice{15.00}
\begin{document}

\copyrightyear{2019}
\acmYear{2019}
\setcopyright{acmlicensed}
\acmConference[DEEM'30]{International Workshop on Data Management for End-to-End Machine Learning}{June 30, 2019}{Amsterdam, Netherlands}
\acmBooktitle{International Workshop on Data Management for End-to-End Machine Learning (DEEM'30), June 30, 2019, Amsterdam, Netherlands}
\acmPrice{15.00}
\acmDOI{10.1145/3329486.3329491}
\acmISBN{978-1-4503-6797-4/19/06}

\title{\sysname: Practical Domain Adaptation with Loss Reweighting}

\author{Justin Chen, Edward Gan, Kexin Rong, Sahaana Suri, Peter Bailis}
\affiliation{
  \institution{Stanford DAWN Project}
}

\renewcommand{\shortauthors}{J. Chen, E. Gan, K. Rong, S. Suri and P. Bailis}

\begin{abstract}
Domain adaptation provides a powerful set of model training techniques given domain-specific training data and supplemental data with unknown relevance.
The techniques are useful when users need to develop models with data from
varying sources, of varying quality, or from different time ranges.
We build \sysname, a system for practical domain adaptation.
\sysname utilizes \reweighting from~\cite{bendavid2010}, which provides consistently high model accuracy across a variety of datasets in our empirical analysis.
However, \reweighting is sensitive to the choice of a weight hyperparameter that is expensive to tune.
We develop optimizations leveraging unique properties of \reweighting that allow \sysname to output accurate models while improving training time compared to na\"ive hyperparameter search.
\end{abstract}

%
%

\maketitle

\section{Introduction}
The availability of large, labeled training datasets has made the development of data-intensive machine learning models increasingly accessible~\cite{lecun1998mnist,deng2009imagenet,dua2017uci,halevy2009data,sun2017revisitdata}.
However, when developing models for new domains, acquiring sufficient  training data for a user's target setting can be costly or infeasible. 
In these scenarios, supplemental but potentially less relevant datasets can also boost model performance by serving as additional training data.
For instance, while a data scientist at a small online retailer may only have access to a limited number of reviews from their retail portal to construct a sentiment prediction model, they can augment their small training set with larger public datasets such as the Amazon reviews dataset~\cite{mcauley2013ratingreview}.
This scenario in which users want to train models using limited training data from the \emph{target} domain with supplemental data from a \emph{source} domain is a form of transfer learning called \emph{domain adaptation} (\DA)~\cite{daume2006domainstat,pan2010transfersurvey}.

\begin{figure}
\includegraphics[width=\columnwidth]{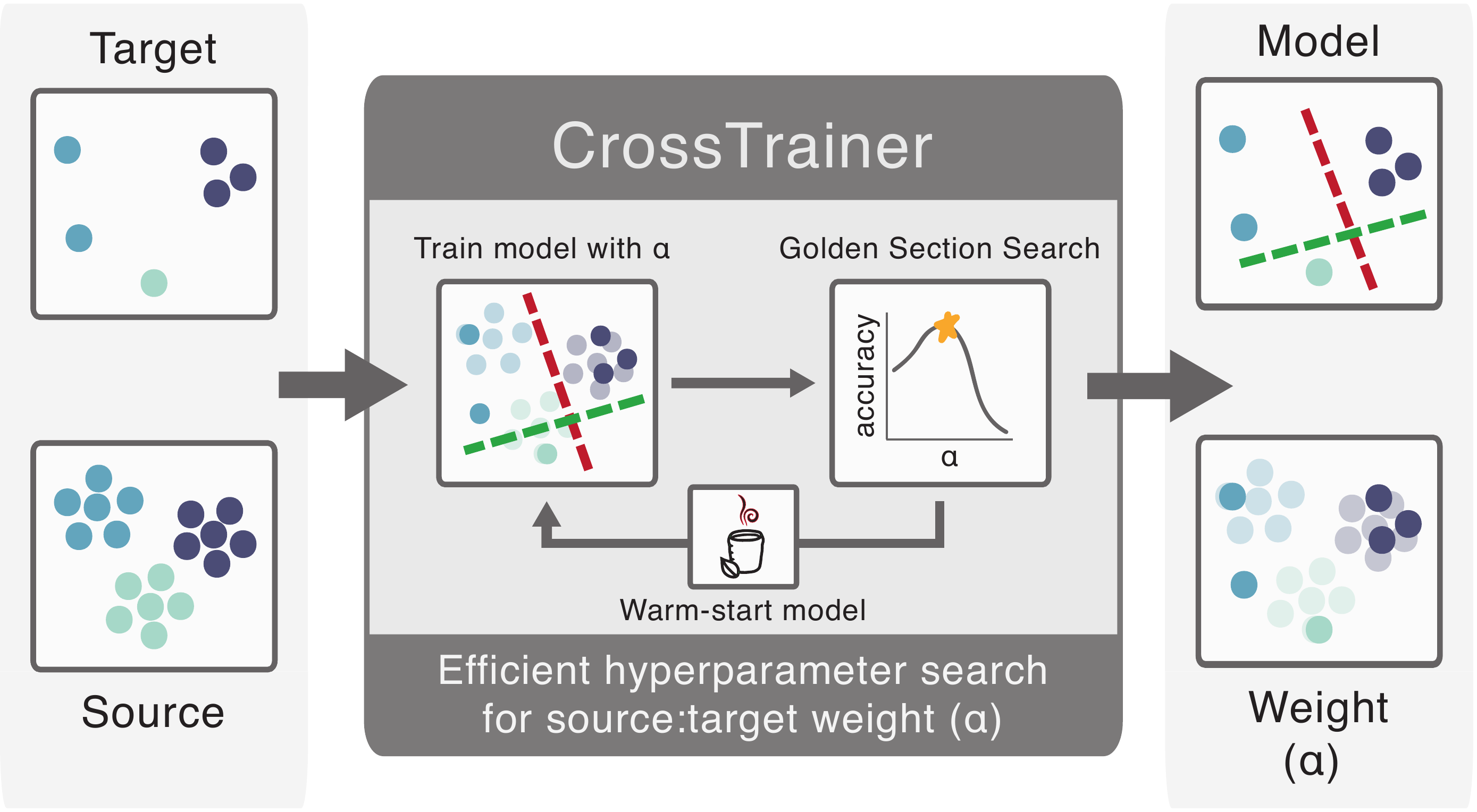}
\caption{\sysname trains a model to maximize validation accuracy by learning a weighting $\alpha$ between target and source data. \sysname addresses the bottleneck of tuning $\alpha$ with optimizations to reduce both the parameter search space and retraining time.}
\label{fig:sys}
\vspace{-1em}
\end{figure}



Despite the common need to deal with data from varying sources and of varying quality, practitioners lack the means to easily apply DA to their machine learning pipelines. In this paper, we develop \sysname, a system for practical DA that integrates with existing ML training workflows, i.e. those using the scikit-learn API. 
\sysname takes source and target datasets as input, and trains a model optimized for accuracy on the target domain (Figure~\ref{fig:sys}). 
In building \sysname, we address two practical challenges that are overlooked in the DA literature. 

First, the machine learning literature proposes a wide range of methods for domain adaptation \cite{daume2007easyda,bendavid2010,dai2007boosting,sun2016coral}, with no clear consensus on their applicability.
Different \DA methods provide different trained model accuracies depending on the target and source datasets, which makes it challenging for users to select the appropriate method a priori.
To address this challenge of method choice, we perform an empirical analysis of six \DA techniques across eight datasets, focusing on techniques that are agnostic to model architecture.
We find that \reweighting~\cite{bendavid2010}, which adjusts the loss function of a base model to balance the relative importance of the target and source datasets, consistently achieves high accuracy across a variety of source and domain types. 
In particular, unlike many alternatives, \reweighting never performs worse than training with target or source data in isolation.
Further, \reweighting is compatible with any base model that uses loss minimization, including logistic regression and gradient boosted decision trees, and requires no adjustments to standard training procedures and feature formats.
As a result, \sysname utilizes \reweighting as its core method to perform domain adaptation. 

However, the accuracy of \reweighting depends heavily on a key hyperparameter $\alpha$ that determines the relative importance of datasets during training. Na\"ively tuning $\alpha$ requires potentially expensive hyperparameter search over many possible values, retraining the model for each configuration. For instance, searching over 100 hyperparameter values for a logistic regression model over 7 million datapoints can take 80 minutes. 

This leads us to a second challenge: \reweighting requires expensive tuning and retraining procedures that can limit its practical application. To address this, \sysname introduces two optimizations for hyperparameter selection that decrease both the number of evaluations and the computational cost of each evaluation. 
First, to prune the search space, we leverage the unimodal relationship between model accuracy and $\alpha$, verified both in practice and in the context of a theoretical upper bound introduced in~\cite{bendavid2010}. As a result, \sysname uses golden section search to prune the hyperparameter space~\cite{gss}. 
Second, we observe that the hyperparameter search consists of repeatedly training models under similar conditions, yielding model coefficients that are closer to each other than random initialization. To decrease the cost of each evaluation of $\alpha$, \sysname utilizes warm-start initialization~\cite{chu2015warm}, further reducing training time for models trained with iterative procedures such as stochastic gradient descent.
Together, these optimizations improve \reweighting performance by up to $48\times$ compared to a grid search over 100 values. 


In summary, we make the following contributions: 
\begin{itemize}[leftmargin=*,align=left,topsep=1ex]
    \item We empirically evaluate six domain adaptation techniques and find that \reweighting provides consistent best or second-best accuracy across real-world and synthetic workloads.
    \item Based on these results, we introduce \sysname, a system utilizing \reweighting to provide accurate and rapid domain adaptation for loss-minimizing models.
    \item We develop optimizations that provide up to $48\times$ speedup for \reweighting compared to na\"ive hyperparameter tuning, and analyze the soundness of the optimizations. 
\end{itemize}

The remainder of this paper proceeds as follows.
In Section~\ref{sec:background}, we introduce our system and its usage.
In Section~\ref{sec:method}, we provide details on the \reweighting technique for domain adaptation.
In Section~\ref{sec:optimizations}, we describe runtime optimizations for \sysname.
In Section~\ref{sec:eval}, we evaluate \sysname for accuracy, robustness, and runtime.
In Section~\ref{sec:related}, we discuss related work.
We conclude in Section~\ref{sec:discussion}.

\section{System Overview}
\label{sec:background}
\sysname is a system for improving machine learning model accuracy using domain adaptation. 
Given input target and source datasets, as well as a model to fit, \sysname trains the model using the \reweighting technique from~\cite{bendavid2010}, optimizing for model validation accuracy on the target domain (Figure~\ref{fig:sys}).
Users can separately configure hyperparameter settings for model training and \reweighting. 
The major \sysname parameters are $k$, the number of folds used for cross-validation on the target data, and $\delta$, the desired precision of the hyperparameter tuning procedure for \reweighting's $\alpha$ parameter (i.e., estimate the optimal $\alpha^*$ with respect to validation accuracy within $\alpha^* \pm \delta$).

\begin{figure}
\begin{lstlisting}[language=Python]
from crosstrainer import CrossTrainer
from sklearn import linear_model
lr = linear_model.LogisticRegression(...)
ct = CrossTrainer(lr, k=5, delta=0.01)
lr, alpha = ct.fit(
  X_target, Y_target,  # task target
  X_source, Y_source)  # supplemental source
Y_pred = lr.predict(X_test)
\end{lstlisting}
\caption{\sysname usage with scikit-learn. Here we fit a logistic regression classifier, but other classifiers are also compatible. \label{fig:codeexample}}
\vspace{-1em}
\end{figure}

We implement \sysname in Python\footnote{\url{https://github.com/stanford-futuredata/crosstrainer}}. In Figure~\ref{fig:codeexample}, we demonstrate how \sysname can serve as a drop-in supplement for existing machine learning workflows using the scikit-learn API~\cite{scikit-learn}. 
\sysname does not require any modifications to feature pre-processing or downstream monitoring pipelines, as it trains models that operate on target ($S_T=(X_T,Y_T)$) and source ($S_S=(X_S,Y_S)$) features directly.
As output, \sysname produces a model that performs better than models trained na\"ively on only the target or source data separately.

To emphasize \sysname's broad applicability, we describe three motivating use cases. 

\minihead{Sentiment Analysis} Consider the case of an online shoe retailer who wishes to build a sentiment classifier for reviews on her website.
While she can curate a labeled target dataset specific to her needs, it might be expensive for her to collect and label a large enough dataset~\cite{halevy2009data} to effectively train a model.
However, if the retailer were to make use of a large source dataset of labeled Amazon reviews~\cite{mcauley2013ratingreview} in conjunction with a small amount of target shoe review data, \sysname would enable her to train a model that outperforms those trained on either dataset separately.

\minihead{Data Cleaning} Consider the case of an advertisement content sensor who wants to build a movie genre classifier to detect horror movies using publicly available datasets from IMDb~\cite{imdb} and Yahoo~\cite{yahoo}. The IMDb dataset has 50K examples, but its genre labels are noisy (e.g. having ``Kids" and ``Horror" tagged for the same movie). In contrast, the Yahoo dataset contains only 10K examples but is much cleaner. Correct handling of dirty data has the potential to significantly improve model performance~\cite{krishnan2016activeclean}. In this case, \sysname can help users balance the two related datasets and improve the final classifier performance. 

\minihead{Time-varying behaviors} Consider the case of an airline operator who would like to train a model for predicting flight delay status. 
While current flight information is more relevant for training, there might only be a small amount of recent flight data collected and labeled for use. 
With \sysname, the operator can leverage larger datasets with labeled historical data to augment the training and enable accurate prediction on current data. 

\section{\reweighting}
\label{sec:method}
We describe domain adaptation, \reweighting, and \reweighting's properties that enable our optimizations. 

\subsection{Domain Adaptation}
In domain adaptation, a \emph{domain} is pair $(D,f)$ consisting of a distribution $\mathcal{D}$ over inputs $x \in \mathcal{X}$ and a labeling function $f: \mathcal{X} \to \{0,1\}$. 
Data is drawn from a \emph{target} domain $(\mathcal{D}_T, f_T)$ and a related, not always identical \emph{source} domain $(\mathcal{D}_S, f_S)$.
Both domains share the same input space $\mathcal{X}$.
The goal of domain adaptation is to learn a classifier $h$ that minimizes the target error, defined as the probability that $h$ disagrees with a labeling function $f_T$ according to the distribution $\mathcal{D}_T$: 
\begin{equation*}
    \epsilon_T(h) = \mathbb{E}_{x \sim \mathcal{D}_T}[|h(x) - f_T(x)|], \quad h,f_T: \mathcal{X} \to \{0, 1\}.
\end{equation*}
We assume that for training, we have access to $m$ instances, where $\beta m$ instances are drawn from $\mathcal{D}_T$, and $(1-\beta)m$ are drawn from $\mathcal{D}_S$. 

\subsection{Loss Reweighting}
\Reweighting~\cite{bendavid2010} is an instance-based method for supervised domain adaptation that adjusts the loss function of a classifier to weight the relative importance of target and source datasets. 
Loss reweighting seeks to find a classifier $h$ that minimizes the empirical $\alpha$-error $\hat{\epsilon}_\alpha(h)$, which is a linear combination of the empirical source error $\hat{\epsilon}_S(h)$ and empirical target error $\hat{\epsilon}_T(h)$ for a given $\alpha \in [0, 1]$: 
\begin{equation}
\label{eq:a-error}
    \hat{\epsilon}_{\alpha}(h) = \alpha\hat{\epsilon}_T(h) + (1 - \alpha)\hat{\epsilon}_S(h).
\end{equation}
Intuitively, the hyperparameter $\alpha$ provides a means to trade off the importance of the target dataset with the value of a potentially larger but less relevant source dataset. 
By choosing $\alpha$ appropriately, we should be able to learn at least as well as if we only used source ($\alpha=0$) or target data ($\alpha=1$), or if we use the union of the source and target data ($\alpha=\beta$). 

The authors in~\cite{bendavid2010} show that $\hat{\epsilon}_{\alpha}(h)$ is a proxy for the true error on the target distribution $\epsilon_T(h)$.
Specifically, a classifier $\hat{h} = \min_h\hat{\epsilon}_{\alpha}(h)$ that minimizes empirical $\alpha$-error and a classifier $h_T^* = \min_h \epsilon_T(h)$ that minimizes target error satisfies: 
\begin{align} 
\label{eq:upper}
\epsilon_T(\hat{h}) &\leq \epsilon_T(h_T^*) + g(\alpha) \\
\label{eq:f-alpha}
g(\alpha) &= 2B \sqrt{\frac{\alpha^2}{\beta} + \frac{(1-\alpha)^2}{1 - \beta}} + 2(1 - \alpha) A.
\end{align} 
$g(\alpha)$ is thus an upper bound on the cost of optimizing $\hat{\epsilon}_\alpha$ instead of the unknown $\epsilon_T$. 
$A \in \mathbb{R}_{\geq0}$ is a constant measuring the distributional divergence between source and target, and $B \in \mathbb{R}_{\geq0}$ is a constant measuring classifier class complexity, for more details see~\cite{bendavid2010}.

\begin{figure*}[t!]
\centering
\includegraphics[width=0.9\linewidth]{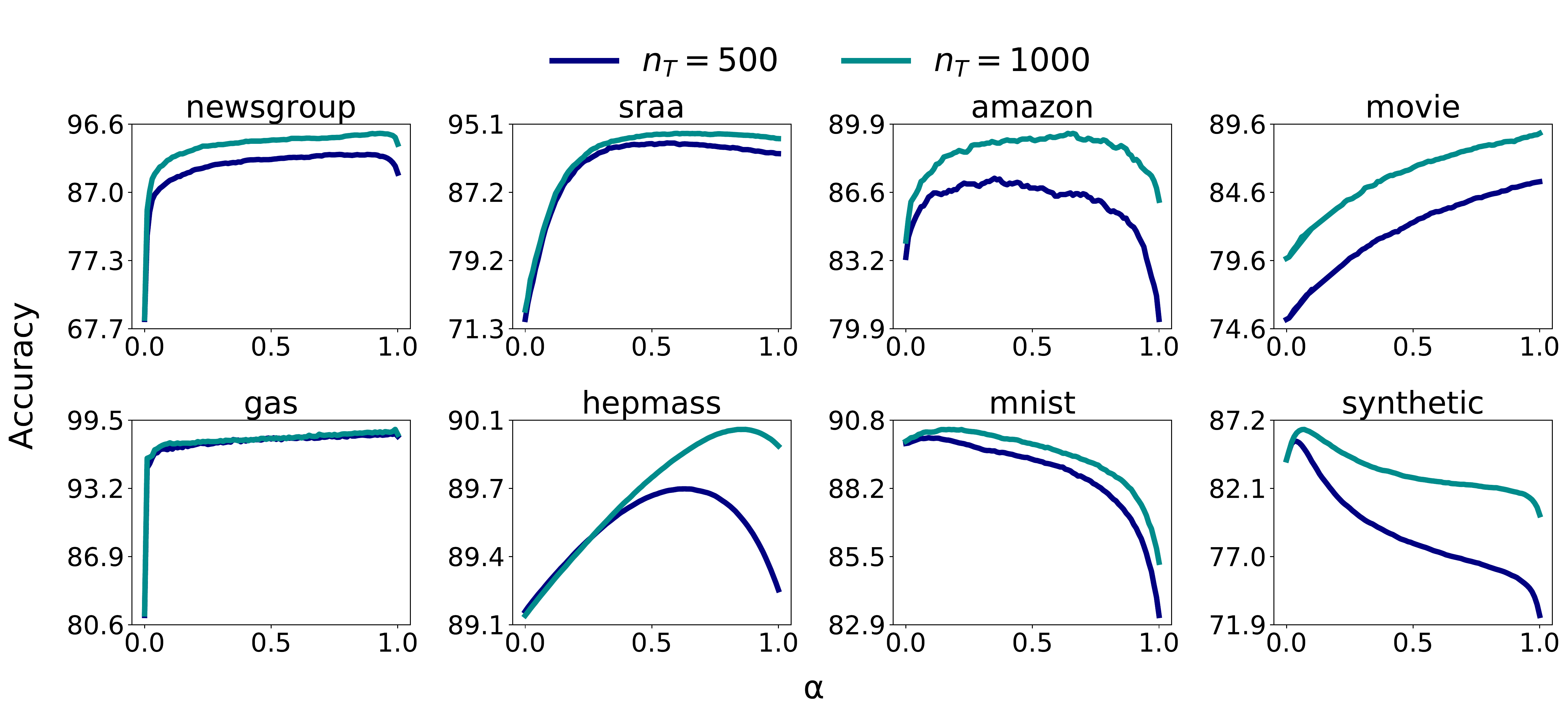}
\caption{Across a variety of datasets, the accuracy curves are unimodal, but not always concave.}
\label{fig:u-shape}
\end{figure*}

\subsection{Properties of $\alpha$ and Loss Reweighting}

In practice, the optimal reweighting parameter $\alpha^*$ depends on both the size and quality of the source and target training sets. 
While one could derive an estimate for $\alpha^*$ from the upper bound in Equation~\ref{eq:f-alpha}, good estimates for $A$ are computationally intractable and optimize for an upper bound rather than the true accuracy. 
Instead, users can empirically tune $\alpha$ as a hyperparameter and optimize for target performance on a validation set. 
However, the repeated training necessary for tuning $\alpha$ are computationally expensive.

In this section we investigate properties of $\alpha$ and $\hat{\epsilon}_\alpha$ that will allow us to reduce the overhead of na\"ive hyperparameter search.
First, we formally verify the empirical observation in~\cite{bendavid2010} that $g(\alpha)$ is convex, and thus there exists a single value $\alpha_g$ that minimizes $g(\alpha)$.

\minihead{Proof of convexity of error upper bound} Let
\begin{equation*}
    z(\alpha) = \sqrt{\frac{\alpha^2}{\beta} + \frac{(1-\alpha)^2}{1 - \beta}}.
\end{equation*}
Then,
\begin{equation*}
    g''(\alpha) = 2B\cdot z''(\alpha) = - \frac{2B}{(\beta(2\alpha-1)-\alpha^2)\cdot z(\alpha)}.
\end{equation*}
We want to show that $g''(\alpha) \geq 0$.
Since $B \geq 0, z(\alpha) \geq 0$, we must show that $\beta(2\alpha-1)-\alpha^2 \leq 0$. There are two cases: 
\begin{enumerate}[itemsep=8pt]
    \item $\alpha \leq 0.5$: We have $(2\alpha -1) \leq 2\times0.5 -1 = 0$, also given $\beta \geq 0, -\alpha^2 \leq 0 \Rightarrow \beta(2\alpha-1)-\alpha^2 \leq 0$.
    \item $\alpha > 0.5$:
    Since $\beta \in [0, 1]$, $\beta(2\alpha-1) \leq 2\alpha - 1$. Therefore
    \[\beta(2\alpha-1)-\alpha^2 \leq 2\alpha -1 - \alpha^2 = -(\alpha-1)^2 \leq 0. \] 
\end{enumerate}
Therefore, $g(\alpha)$ is convex.

\minihead{Empirical evidence} Equation~\ref{eq:f-alpha} shows that the error is upper bounded by a convex function of $\alpha$, but we must verify that the empirical error curve itself is convex in practice. 
The authors in~\cite{bendavid2010} show a few promising examples, and we extend their empirical evaluation to a wider range of datasets.

We train a series of logistic regression models that minimize empirical $\alpha$-error under varying $\alpha$. 
Figure~\ref{fig:u-shape} reports the achieved test accuracy for two different target dataset sizes on a number of datasets. As we report accuracy instead of error, we expect the curves to be concave, not convex.

We first observe that for a wide range of target and source datasets, the optimal choice of $\alpha^*$ is not at $\alpha=0$ or $\alpha=1$, indicating that the identification of $\alpha^*$ requires tuning.
Further, contrary to the observations in~\cite{bendavid2010}, the empirical target accuracy is sometimes not concave (Figure~\ref{fig:u-shape}). 
However, the accuracy curves are still all \emph{unimodal}, a weaker property than concavity that describes functions with a unique local maximum. 
A function $f(x)$ is defined as unimodal if there exists a unique value $x^*$ that maximizes the function, and that the function monotonically increases for $x \leq x^*$ and monotonically decreases for $ x > x^*$.
Unimodality is sufficient to enable optimizations in Section~\ref{sec:gss} that improve hyperparameter search performance.
Although for some datasets, the results have small fluctuations (Figure~\ref{fig:u-shape}) and are not strictly unimodal, in practice these do not hinder the effectiveness of our optimizations.
\section{Optimizing \sysname}
\label{sec:optimizations}

To enable efficient usage of \reweighting in \sysname, we reduce the computational overhead of tuning the $\alpha$ parameter.
We introduce two optimizations: we reduce the range of $\alpha$ to evaluate using golden section search, and reduce the runtime for each iteration using warm start. 

\subsection{Golden Section Search} 
\label{sec:gss}
Hyperparameter search is a well studied problem with simple solutions such as grid search, to state of the art improvements to random search such as Hyperband~\cite{li2016hyperband} and Bayesian optimization~\cite{klein2016fast}.
However, unlike in traditional, domain-agnostic hyperparameter search, here we can take advantage of the unimodal relationship between $\epsilon_T(\hat{h})$ and $\alpha$ to more quickly hone in on $\alpha^*$.

As we wish to approximate the optimal $\alpha^*$ to a precision of $\delta$, we begin by considering grid search with a granularity of $\delta$ (i.e., with $\frac{1}{\delta}$ grid points) as a baseline. 
However, for logistic regression on datasets with a few million points and with $\delta = 0.01$, the grid search procedure can take on the order of an hour on a modern processor. 

\begin{figure} 
\centering
\includegraphics[width=0.75\linewidth]{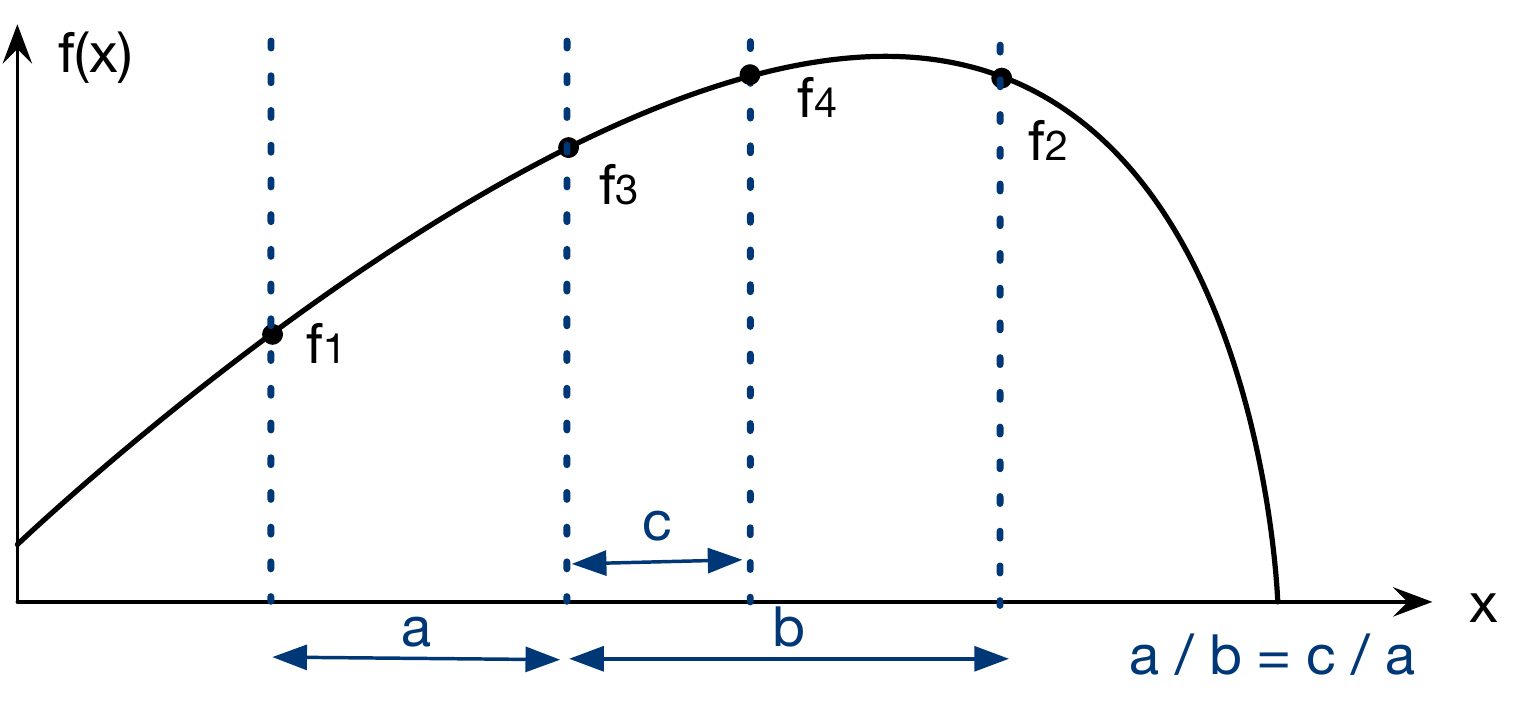}
\caption{Illustration of golden section search. The spacing between probe points are iteratively narrowed according to the golden ratio. }
\label{fig:gss}
\vspace{-1em}
\end{figure}

Concretely, unimodality allows us to use golden section search (Figure~\ref{fig:gss}) to find a value of $\alpha$ that minimizes target error \cite{gss}. 
Just as binary search reduces the search space for finding specific values of a monotonic function, golden section search reduces the search space for locating the extremum of a unimodal function.
The algorithm works by iteratively narrowing the range of a bracket of three probe points that covers the location of the extremum. 
Each narrowing reduces the range of the probe points by a factor of $1 - \phi^{-1}$ (where $\phi$ is the golden ratio), so, to get brackets with range less than $\delta$, we need $O(\log\left(\frac{1}{\delta}\right))$ iterations, compared to $O\left(\frac{1}{\delta}\right)$ for grid search. 
We find in practice (Table~\ref{tab:comparison}) that minor deviations from unimodality do not affect golden section search's ability to approximate $\alpha^*$.
We provide the pseudocode for hyperparameter tuning $\alpha$ using golden section search in Algorithm~\ref{alg:gss}.
Note that before the golden section search, we first use a form of binary search to find three values of $\alpha$ that must bracket the optimum value.

\begin{algorithm}
\small
\caption{Find reweighting factor}
\label{alg:gss}
\begin{algorithmic}[1]
\Function{Find\_Bracket}{l, r, $\delta$}
  \State m = (l + r) / 2
  \State $a_l$ = \Call{Accuracy}{l}, $a_m$ = \Call{Accuracy}{m}, $a_r$ = \Call{Accuracy}{r}
  \If {r - l $< \delta$ \OR \Call{max}{$a_l$, $a_m$, $a_r$} == $a_m$ } 
    \State \Return l, m, r \Comment {Stopping criteria, or bracket found}
    \ElsIf{$a_l \leq a_r$} \Comment{Binary search}
        \State \Return \Call{Find\_Bracket}{m, r}
    \Else
        \State \Return \Call{Find\_Bracket}{l, m}
  \EndIf
\EndFunction

\State
\Procedure{FindWeighting}{$\delta$}
\State left, mid,  right = \Call{Find\_Bracket}{0, 1, $\delta$}
\If {\Call{Accuracy}{right} > \Call{Accuracy}{mid}}
\State \Return right 
\ElsIf {\Call{Accuracy}{left} > \Call{Accuracy}{mid}} 
\State \Return left 
\Else
\State \Return \Call{GSS}{left, mid, right} \Comment{Golden section search}
\EndIf
\EndProcedure
\end{algorithmic}
\end{algorithm}

\subsection{Warm Start}
While golden section search reduces the search space for $\alpha$, we can go further by taking advantage of the fact that each retraining operation is relatively similar to the previous operations, with a minor change in the value of $\alpha$.
When working with models trained iteratively (i.e., using stochastic gradient descent~\cite{robbins1951sgd}, or SGD), we can take advantage of warm start optimizations to reduce the cost of each successive retraining.
Intuitively, as we train on the same data with slightly different objective functions, the feature weights should remain similar across different choices of $\alpha$. 

Thus, in \sysname when optimizing SGD-based classifiers, we save the trained model coefficients for different $\alpha$ during golden section search and re-use them to initialize SGD when training for new values of $\alpha$. This reduces the number of SGD epochs required to converge.

While this initial warm start implementation is straightforward, we find it provides runtime improvements in practice. 
More sophisticated warm start techniques for incremental learning ~\cite{tsai2014incremental} and hyperparameter estimation~\cite{chu2015warm} can also be incorporated in future work.
\section{Evaluation}
\label{sec:eval}

We empirically evaluate \sysname in terms of model accuracy and training time. We demonstrate:

\begin{enumerate}[label=\textbf{\arabic*})]
    \item \sysname consistently produces models that match or exceed the accuracy of competing methods on real datasets.
    \item \sysname's performance is robust over a variety of source and target data distributions.
    \item \sysname's optimizations decrease training time while maintaining model accuracy.
\end{enumerate}

\subsection{Experimental Setup}

\minihead{Implementation} We implement \sysname in Python. For most experiments, we use scikit-learn's implementation of logistic regression with stochastic gradient descent~\cite{scikit-learn} as the base model. We also evaluate \sysname with gradient boosted decision trees (GBDTs), using the XGBoost package~\cite{chen2016xgboost}.
By default, we use $k$-fold cross-validation with $k=5$ to optimize model- and method-specific hyperparameters including the regularization parameter for logistic regression, the number of estimators (trees) for GBDTs, and $\alpha$ for \sysname. 

\minihead{Baselines} We compare against three standard domain adaptation baselines:

\begin{enumerate}[leftmargin=*,align=left]
    \item \textit{Target}: This baseline trains a single model on only the target data ($\alpha = 1$).
    \item \textit{Source}: This baseline trains a single model on only the source data ($\alpha = 0$).
    \item \textit{All}: This baseline trains on uniformly weighted union of the source and target datasets ($\alpha = \beta$).
\end{enumerate}

\minihead{Competing Methods} In addition to the baselines, we evaluate the following domain adaptation methods.  

\begin{enumerate}[leftmargin=*,align=left]
\item \textit{\sysname}: Our implementation of \reweighting including optimizations. Unless otherwise specified, we use $\delta = 0.01$ throughout the experiments.
\item \textit{Pred}: This method trains a class-balanced logistic regression model to distinguish between the source and target training examples. The trained model weights each instance of the combined dataset by the predicted probability of the instance belonging to the target dataset. 
\item \textit{Import}: Similar to Pred, this method trains a logistic regression model to distinguish between source and target for instance-specific reweighting. Given a predicted probability $p=\Pr(\text{Target}|x,f(x))$, the instance $x$ is weighted by $w=c/\left(\frac{1}{p}-1\right)$ where $c = \frac{n_S}{n_T}$, an asymptotically optimal importance weighting for covariate shift~\cite{shimodaira2000covariateshift}. 
\item \textit{FeatAug \cite{daume2007easyda}}: This method first augments input features of the source and target data, and trains a model on the combination of the augmented data. Given original features $x$, source features are represented as $\langle x, x, 0\rangle$ and target features are represented as $\langle x, 0, x\rangle$.
\item \textit{TrAdaBoost \cite{dai2007boosting}}: This method uses a variant of boosting by iteratively training an ensemble of models with misclassified target instances up-weighted, and misclassified source instances down-weighted at each iteration. Note that TrAdaBoost is only implemented for binary classification tasks.
\item \textit{CORAL \cite{sun2016coral}}: This method aligns second-order statistics of the source and target datasets as a pre-processing step before training on the combined datasets. CORAL can also be used for semi-supervised domain adaptation, but we apply it here in a supervised setting.
\end{enumerate}

\minihead{Datasets}  We perform evaluation on 7 real-world datasets and 1 synthetic dataset (Table~\ref{tab:datasets}).
Of the target data, we set aside $80\%$ for training, and use the remaining $20\%$ as a test set.
As \sysname focuses on regimes with very limited target data, we downsample the target training data to assess performance on small target datasets. 

In addition to the datasets that are commonly used in DA studies (Newsgroups~\cite{newsgroups}, SRAA~\cite{sraa} and Amazon reviews~\cite{blitzer2007amazon}), we evaluate on the IMDb~\cite{imdb}/Yahoo~\cite{yahoo} movie datasets, UCI Gas Sensor Array Drift \cite{dua2017uci, vergara2011gas}, the UCI HEPMASS \cite{dua2017uci, baldi2016hepmass}, and the MNIST \cite{lecun1998mnist,loosli2006mnistinifinite} datasets. 
We describe how each dataset is split into target and source in the Appendix~\ref{sec:appx}. 

We construct the synthetic dataset as follows.
Both source and target are sampled from the same Gaussian distribution in 500 dimensions. 
The labeling function for target data is a zero-one threshold function applied to a linear function of the input features given by $g(x) = \sum_{j=1}^{500} x_j$, plus Gaussian ($\mathcal{N}(0,1)$) noise. The labeling function for the source data is equivalent except the weights of the linear component are changed to $g(x) = \sum_{j=1}^{500} c_j x_j$, where $c_j \sim \mathcal{N}(1,\sigma)$.

\begin{table}
  \caption{Datasets and classification tasks used in the evaluation.}
  \label{tab:datasets}
  \small
  \begin{tabular}{lccrrr}
    \toprule
    Name & Clf Task & Domain Split & $n_{S}$ & $n_{T}$ & $d$\\
    \midrule
    Newsgroups & Topic   & Category    & 3.5K & 3.6K  & 50K \\
    SRAA       & Topic   & Category    & 8K   & 8K & 60K \\
    Amazon     & Review  & Product         & 2K  & 2K & 500K \\
    movie      & Genre  & Site Source     & 80K  & 10K   & 100 \\
    gas        & Chem   & Time     & 5.9K & 4.4K & 128 \\ 
    hepmass    & Binary  & Particle   & 7M   & 7M & 26 \\
    MNIST      & Image  & Perturbation & 70K & 60K & 784 \\
    synthetic  & Binary & Label Shift     & 100K & 100K  & 500 \\ 
  \bottomrule
\end{tabular}

\end{table}

\subsection{Empirical Evaluation of DA Methods}
\label{sec:reweighting_eval}

\begin{table*}
\small
  \caption{Comparison of output model accuracy. \sysname outperforms related methods on most datasets and always performs at least as well as the baselines. We report the performance of \reweighting with no optimizations along with \sysname. Bold entries achieved accuracy within 0.1\% of the best method and starred entries did not underperform any of the baselines.}
  \label{tab:comparison}
  \begin{tabular}{ll|lllllllll}
    \toprule
    Dataset & $n_{T}$ & Target & Source & All & \sysname (unopt.) & Pred & Import &  FeatAug & TrAdaBoost & CORAL \\
    \midrule
    Newsgroups & 500 & 90.1 & 68.6 & 88.9 & \textbf{91.9 (91.7)*} & 90.4* & 89.3 & 90.3* & 85.3  & 70.1\\
    Newsgroups & 1000 & 93.5 & 68.6 & 92.3 & \textbf{95.3 (95.2)*} & 94.0* & 93.3 & 93.8* & 89.1 & 74.1 \\
    SRAA & 500 & 91.6 & 77.1 & 85.8 & \textbf{92.5 (92.5)*} & 90.3 & 85.5 & 90.1 & 89.1  &  63.7 \\
    SRAA & 1000 & 93.3 & 77.1 & 89.1 & \textbf{94.0 (93.9)*} & 92.6 & 89.0 & 92.4 & 90.7 & 69.2 \\
    Amazon (b $\rightarrow$ e) & 500  & 79.2 & 70.8 & 78.6 & 79.2 (79.2)* & \textbf{80.3*} & \textbf{80.3*} & 78.6 & 77.6 & 74.3 \\
    Amazon (b $\rightarrow$ e) & 1000 & 82.4 & 70.8 & \textbf{83.7} & \textbf{83.7 (83.7)*} & 82.7 & 83.0 & 83.0 & 82.1 & 75.8\\
    Amazon (e $\rightarrow$ k) & 500  & 80.9 & 84.4 & \textbf{86.8} & \textbf{86.8 (86.9)*} & 84.2 & 85.0 & 83.2 & 85.5 & 79.6 \\
    Amazon (e $\rightarrow$ k) & 1000 & 85.7 & 84.4 & 88.6 & \textbf{89.5 (89.5)*} & 87.4 & 87.8 & 87.6 & 87.6 & 81.1\\
    movie & 500 & 85.1 & 79.7 & 85.0 & 85.2 (85.1)* & 86.3* & 85.9* & \textbf{87.2*} & 86.2* & 82.1 \\
    movie & 1000 & \textbf{88.8} & 79.7 & 87.0 & \textbf{88.9 (88.9)*} & 87.0 & 85.7 & 88.6 & 86.6 & 86.1 \\
    gas & 500 & \textbf{98.4} & 81.4 & 96.7 & \textbf{98.5 (98.5)*} & 98.1 & 87.3 & 98.0 & -- & 79.3  \\
    gas & 1000 & \textbf{98.7} & 81.4 & 97.6 & \textbf{98.7 (98.7)*} & 98.2 & 90.1 & 98.6 & -- & 83.9  \\
    hepmass & 500 & 89.2 & 89.1 & 89.1 & 89.7 (89.7)* & 89.8* & 89.1 & 88.8 & \textbf{89.9*} & \textbf{90*} \\
    hepmass & 1000 & \textbf{89.9} & 89.1 & 89.1 & \textbf{90.0 (90.0)*} & 89.7 & 89.1 & 88.8 & \textbf{89.9*} & \textbf{90*} \\
    MNIST & 500  & 83.3 & 89.5 & 89.5 & \textbf{89.9 (89.9)*} & 86.7 & 80.4 & 85.2 & -- & 75.1 \\
    MNIST & 1000 & 85.2 & 89.5 & 89.5 & \textbf{90.0 (90.0)*} & 88.4 & 83.9 & 86.5 & -- & 77.2 \\
    synthetic & 500 & 72.8 & 84.2 & 84.5 & \textbf{85.7 (85.7)*} & 74.0 & 75.6 & 73.2 & 76.1 & 82.6  \\
    synthetic & 1000 & 79.8 & 84.2 & 84.8 & \textbf{86.3 (86.3)*} & 80.1 & 81.2 & 80.2 & 70.0 & 84 \\
  \bottomrule
\end{tabular}
\end{table*}

\minihead{Comparison of Target Accuracy} In Table \ref{tab:comparison}, we report the target accuracy of different domain adaption methods and baselines under a total of eighteen settings. We summarize the key observations below.

First, simple baselines are surprisingly difficult to beat; \sysname is the only method that consistently outperforms all baselines. For example, when source-only outperforms target-only (e.g., Amazon (e $\rightarrow$ k) with $n_T = 500$), Pred, Import, FeatAug, TrAdaBoost, and CORAL all perform worse than simply training on the union of the source and target data. 
However, since \reweighting tunes $\alpha$ using the validation set, it is able to adapt to circumstances where the source data is more or less useful compared to the target.

Second, \sysname works across different types of domain splits. For datasets such as MNIST, the distribution of labels $p(y),$ $p(y|x)$ is similar between target and source, while the distribution of the covariates $p(x)$ is different as the infinite MNIST~\cite{loosli2006mnistinifinite} (source) features are generated from deformations and translations of the MNIST (target) features. For datasets like Synthetic, the covariate distributions are equivalent with the difference coming from the labeling distribution $p(y|x)$. In both situations, \sysname significantly outperforms other methods.

As a whole, \sysname using \reweighting is able to produce high-quality models, and achieves the best target accuracy in 15 out of the 18 evaluated scenarios.
To better understand the robustness and generality of these results, we further vary a number of parameters in our experiments including dataset size, the discrepancy between source and target, and the types of models.

\begin{figure}
\caption{Comparison of target accuracy over varying sizes of the target dataset for SRAA (top) and MNIST (bottom).}
\includegraphics[width=0.9\columnwidth]{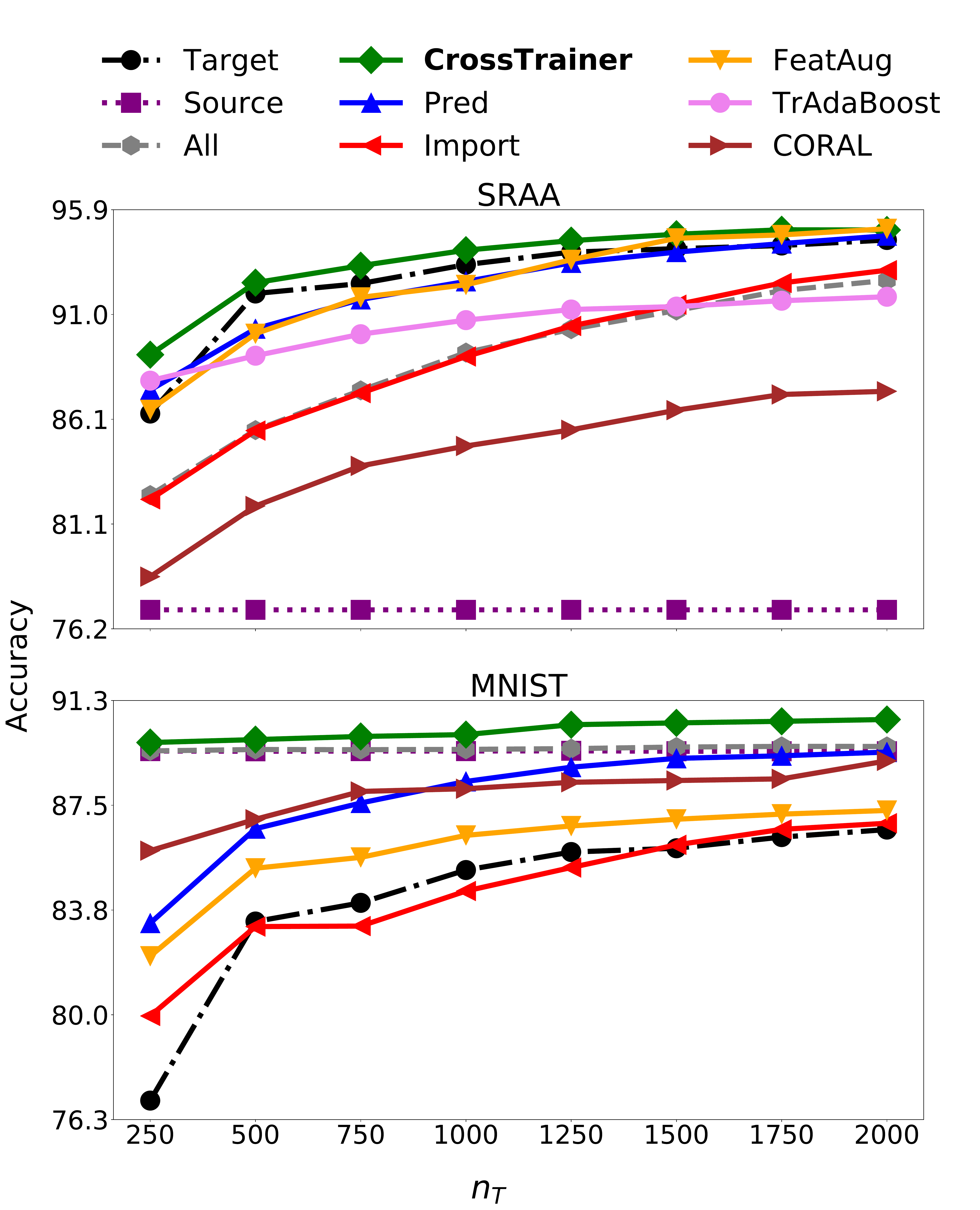}
\label{fig:size}
\end{figure}




\minihead{Robustness to target dataset size} 
In Figure~\ref{fig:size}, we evaluate the effect of target dataset sizes on the performance of different domain adaption methods on the SRAA and MNIST datasets.
Across both datasets and all methods, performance tends to increase with more training data from the target domain.
For the SRAA dataset, with small amounts of target data (Figure~\ref{fig:size}, SRAA, $n_T < 1250$), all domain adaptation methods except for \sysname perform worse than the target-only baseline.
On the other hand, we observe that source data helps improve model performances over the target-only baseline for most methods on the MNIST dataset. However, with small amounts of target data (Figure~\ref{fig:size}, MNIST, $n_T < 1500$), all methods except for \sysname perform worse than the source-only or all baselines.
For both datasets, \sysname consistently outperforms baselines and achieves the best accuracy across the board. 


\begin{figure}
\caption{Comparison of output model accuracy over varying the difference between the source and target distributions for Synthetic with $n_{T} = 500$. \\}
\includegraphics[width=0.85\columnwidth]{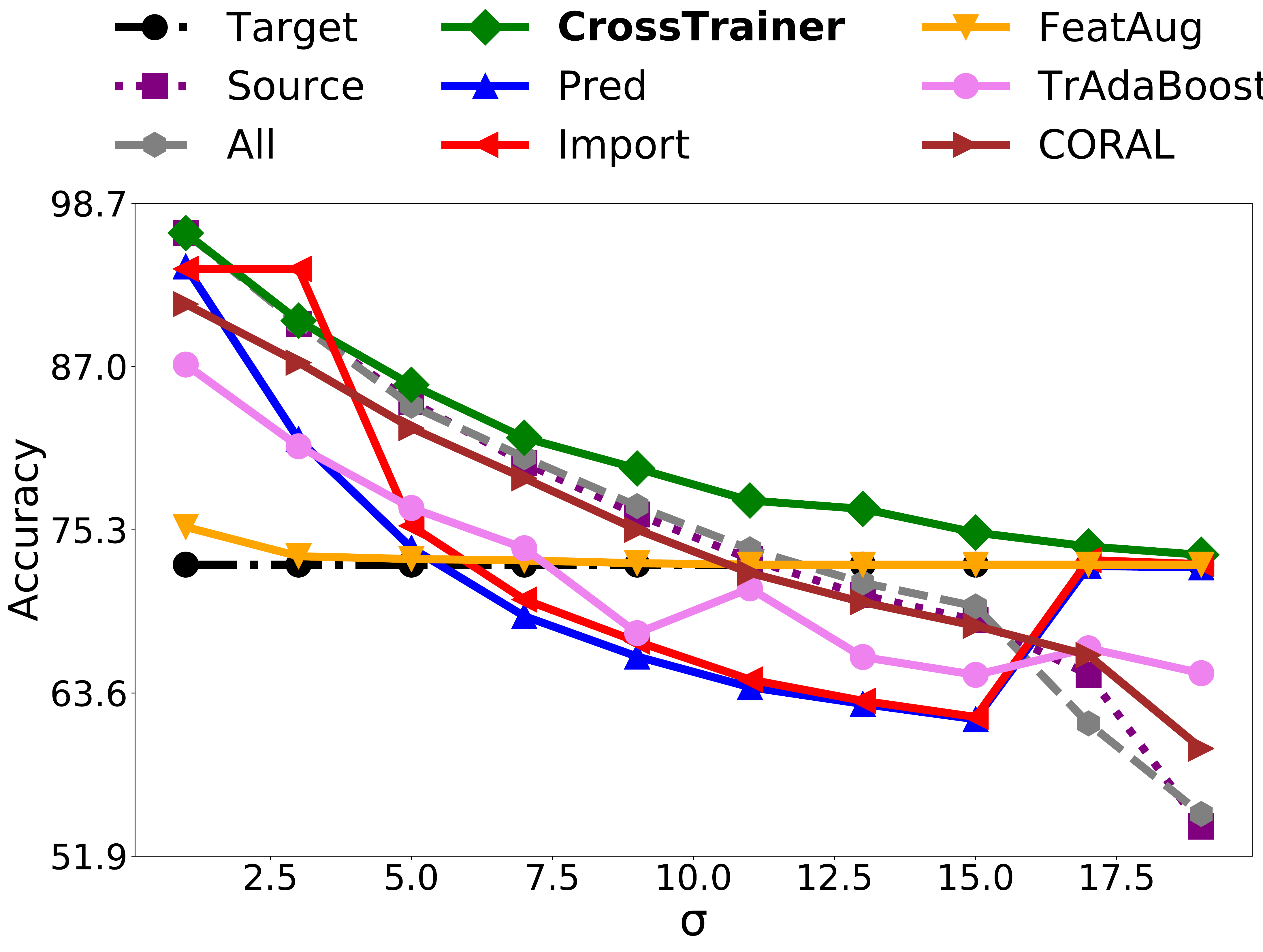}
\label{fig:synthetic-source}
\end{figure}

\minihead{Robustness to source distribution change} In Figure \ref{fig:synthetic-source}, we evaluate \sysname and other DA methods subject to changes to the source distribution on the synthetic dataset. We vary $\sigma$, the parameter that defines the difference between the target and source labeling distributions. Larger values of $\sigma$ indicate that the source distribution is further away from the target distribution. When the source distribution is very close to the target distribution, \sysname matches the performance of the source-only model. At the other extreme, when the source distribution is very different from the target distribution, \sysname matches the performance of the target-only model. Between these extremes, \sysname outperforms all methods except Import when $\sigma = 3$. 

\minihead{Robustness to model type} In Table \ref{tab:gbdt}, we compare \\ \sysname to baselines and competing methods using gradient boosted decision trees (GBDTs) for three datasets with $n_T = 1000$. As GBDTs are not trained using SGD, the warm-start optimization does not apply, but \sysname is still able to use golden section search. 
As with logistic regression, we find that \sysname outperforms competing methods and does not do worse than the baselines.

{
\setlength{\tabcolsep}{3pt}
\begin{table}
\footnotesize
  \caption{Domain adaptation using GBDT models with $n_T = 1000$. Bold entries achieved accuracy within 0.1 of the best method and starred entries did not underperform any of the baselines.}
  \label{tab:gbdt}
  \begin{tabular}{lllllllll}
    \toprule
    Dataset & Target & Source & All & \sysname & Pred & Imp. &  FeatAug \\
    \midrule
    Newsgroups & 84.4 & 62.3 & 79.5 & \textbf{85.1*} & 84.2 & 81.7 & 82.6  \\
    SRAA & 91.7 & 84.8 & 90.1 & \textbf{92.2*} & 91.7 & 90.1 & 92.0*  \\
    MNIST & 86.4 & 92.5 & \textbf{93.4} & \textbf{93.4*} & \textbf{93.3} & \textbf{93.3} & 92.9 \\
  \bottomrule
\end{tabular}
\end{table}
}

\subsection{Effect of Optimizations}
\begin{figure}
    \centering
    \caption{Comparison of search strategies for $\alpha$ with $n_T=1000$. CrossTrainer utilizes golden section search to quickly converge to a value of $\alpha$ that yields high accuracy. }
    \label{fig:search}
    \includegraphics[width=0.9\linewidth]{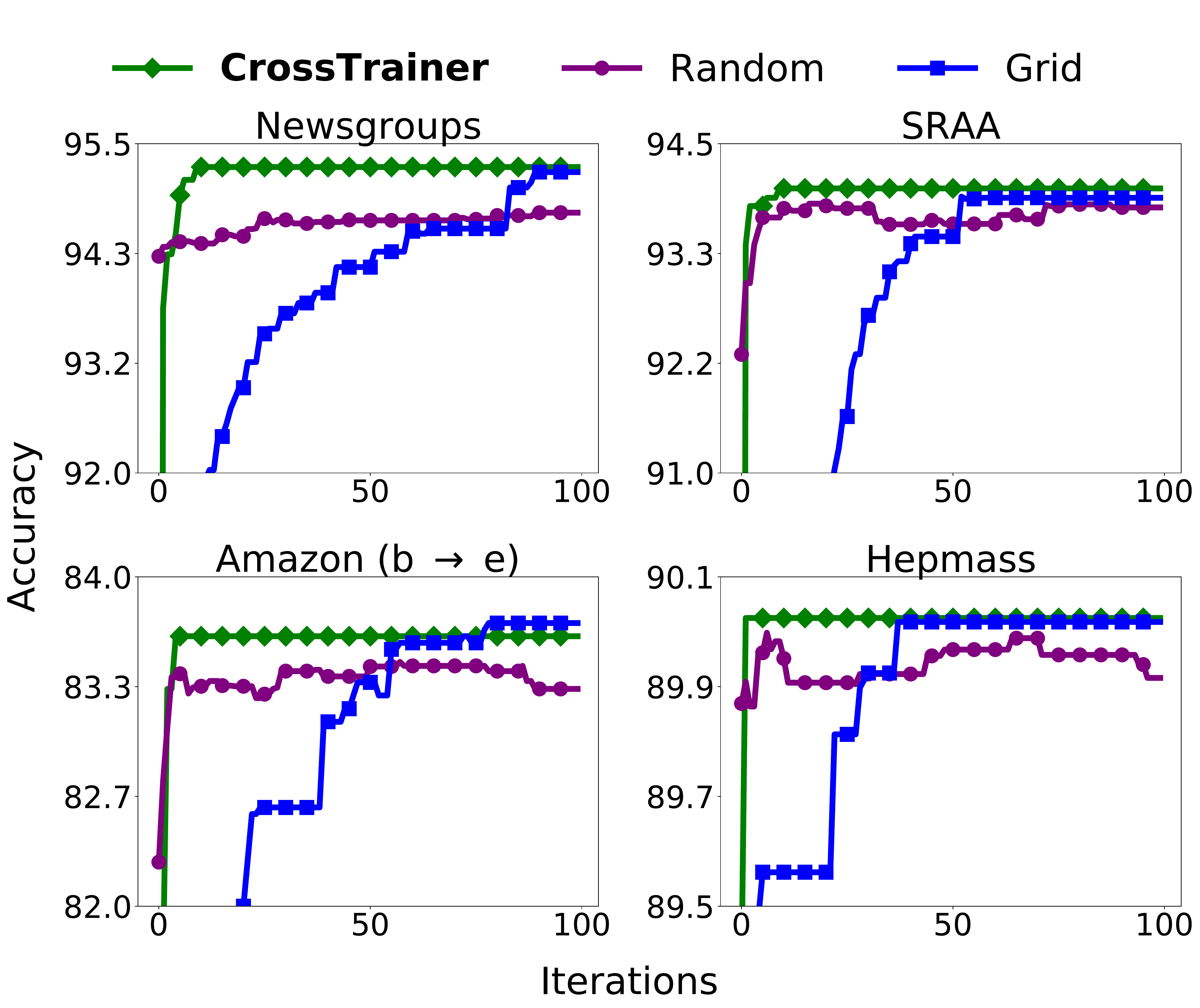}
\end{figure}

\begin{table}
  \small
  \caption{Cumulative factor analysis of total training time for loss reweighting with $\delta = 0.01$. \sysname provides speedups between $11-48\times$ over unoptimized loss reweighting.}
  \label{tab:optimizations}
  \begin{tabular}{lllll}
    \toprule
    Dataset &  Baseline & +GSS & +Warm-start & Speed-up\\
    \midrule
    Newsgroups & 15.5s  & 2.1s  & 1.1s  & 15$\times$ \\
    SRAA      & 7.2s   & 1.0s  & 0.7s & 11$\times$ \\
    Amazon    & 14.2s  & 2.5s  & 1.0s  & 14$\times$ \\
    movie     & 2.2m   & 15.8s & 5.3s  & 25$\times$ \\
    gas       & 18.5s  & 2.6s  & 0.7s & 26$\times$ \\
    hepmass   & 80.1m  & 8.9m  & 1.7m  & 48$\times$ \\
    MNIST     & 31.4m  & 5.1m  & 1.7m & 19$\times$ \\
    synthetic & 20.8m  & 2.2m  & 38.3s & 32$\times$ \\
  \bottomrule
\end{tabular}
\end{table}

We first evaluate the effectiveness of golden section search as a hyperparameter search strategy. 
Figure~\ref{fig:search} compares the target accuracy achieved by golden section search, grid search and random search after a fixed number of search iterations.
We find that across multiple datasets, golden section search converges within at most fifteen iterations.
Random search can reach high accuracies very quickly, but it often fails to match the final performance of golden section search even with 100 iterations.
Grid search eventually achieves high accuracies, but takes many more iterations to converge than golden section search.

We also evaluate the end-to-end performance gains of our optimizations. Table~\ref{tab:optimizations} shows that golden section search and warm-start initialization decrease training time by up 11-$48\times$ compared to the baseline (grid search without warm starts).
Golden section search consistently provides an order of magnitude improvement in training time over grid search.
Warm-starting improves training time in each case, with more of an effect for models that take longer to run, offering up to 5$\times$ speedup.

Finally, we validate the soundness of our optimizations by verifying that they do not reduce the final accuracy of the models trained. 
We compare test set accuracy for \sysname with optimizations enabled with those from a baseline grid search of 100 values (unopt.) in Table~\ref{tab:comparison} and find that our optimizations have a very small impact on accuracy: the difference is consistently less than the differences between alternative methods.

\section{Related Work}
\label{sec:related}

We develop a system for domain adaptation, a type of transfer learning where one has different but related data domains, but the same underlying feature space and tasks \cite{daume2007easyda,bendavid2010,pan2010transfersurvey}. 
We focus on supervised settings for domain adaptation as opposed to the unsupervised case where the learner only has access to an unlabeled target dataset~\cite{sun2016coral}.

Different techniques for domain adaptation can be categorized as in \cite{pan2010transfersurvey,weiss2016survey,deeptransfersurvey}: 
\begin{enumerate}[itemsep=2pt,topsep=1pt]
    \item \emph{Instance-based} methods that integrate source and target instances by assigning a weight to each instance based on its estimated relevance. Examples include TrAdaBoost~\cite{dai2007boosting} and loss reweighting~\cite{bendavid2010}.
    \item \emph{Feature-based} methods that align features from different domains by either transforming features or by learning common feature structures. Examples include feature augmentation~\cite{daume2007easyda} and CORAL~\cite{sun2016coral}. 
    \item \emph{Parameter-based} methods that share parameters from models or priors between the source and the target domain, including many methods from deep transfer learning. Examples include~\cite{yosinski2014transferable,tlsharednn,tommasi2010safety}.
\end{enumerate}
In practice, instance-based methods can be applied as a drop-in wrapper to any training pipeline that supports weighting input data.
This makes instance-based methods a natural choice for building a generic system.
In contrast, feature-based methods may require modifying feature preprocessing and analysis routines, and parameter-based methods are most effective when developed and tuned for specific model architectures (i.e. deep neural nets). 

Thus, in this paper, we primarily compare against instance-based methods as well as a number of representative feature-based methods.
The ``frustratingly easy'' domain adaptation technique in~\cite{daume2007easyda} is also designed for ease-of-use and speed, but as acknowledged by the authors, does not provide consistently accurate results in some settings.
Loss reweighting is proposed in \cite{bendavid2010} as a theoretical framework for domain adaptation. 
However, the authors in~\cite{bendavid2010} do not address the accuracy of their method compared to other techniques or consider runtime and usability.

We draw inspiration from other systems for machine learning with limited training data in other settings including data cleaning and weak supervision \cite{snorkel,holoclean,krishnan2016activeclean}.
Additional related settings include semi-supervised learning~\cite{chapelle2010semisupervised}, where one has unlabeled source data from the same domain, covariate shift~\cite{bickel2009covariate}, where one has unlabeled target data and the conditional label assignment is known to be the same between domains, and active learning~\cite{burr2012active}, where one can iteratively acquire new target training examples.
\section{Conclusion}
\label{sec:discussion}
We proposed \sysname and evaluated its performance for accurate, robust, and fast domain adaptation. 
\sysname utilizes loss reweighting, which we find to be consistently effective, and incorporates optimizations to alleviate the overhead of hyperparameter tuning.


Looking forward, potential extensions of \sysname to support additional methods and input types are promising directions for future work. 
The simplicity of the loss reweighting algorithm suggests that it may be helpful to use reweighting simultaneously with other domain adaptation methods, as well as to incorporate cross-validation hyperparameter tuning more deeply into these methods.
Extending the evaluation and optimizations to support simultaneous adaptation from multiple sources would further improve usability in scenarios where users have many potential source domains to draw from.


{
\footnotesize
\begin{acks}
We thank the members of the Stanford InfoLab as well as Sanjay Krishnan for valuable feedback. This research was supported in part by affiliate members and other supporters of the Stanford DAWN project---Ant Financial, Facebook, Google, Intel, Microsoft, NEC, SAP, Teradata, and VMware---as well as Toyota Research Institute, Keysight Technologies, Northrop Grumman, Hitachi, and the NSF Graduate Research Fellowship grant DGE-1656518.
\end{acks}
}

\appendix

\section{Dataset Tasks}
\label{sec:appx}

\minihead{DA Datasets} We evaluate DA methods on three datasets commonly used in previous DA studies: Newsgroups~\cite{newsgroups}, SRAA~\cite{sraa} and Amazon~\cite{blitzer2007amazon}. For Newsgroups, we use the \emph{rec vs. talk} domain split. For SRAA, we use the \emph{auto vs. aviation} domain split. For Amazon, we use the \emph{b $\rightarrow$ e} and \emph{e $\rightarrow$ k} domain splits.

\minihead{Additional Datasets} We evaluate DA methods on four additional real-world datasets: movie (IMBb and Yahoo), Gas, HEPMASS and MNIST. The task of the movie dataset is to classify movies as horror or comedy based on plot summaries, with target coming from Yahoo and source coming from IMDb. The Gas dataset is split according to time with target batches 5-6 and source batches 7-10. The HEPMASS dataset is split with the target dataset containing collisions of particles with 1,000 mass and the source dataset containing all other collisions. For the MNIST dataset, we use the original dataset for target and use samples from the infinite MNIST dataset \cite{loosli2006mnistinifinite} generated via pseudorandom transformations to the original MNIST data for source.

\bibliographystyle{ACM-Reference-Format}
\bibliography{main}


\begin{thebibliography}{40}


\ifx \showCODEN    \undefined \def \showCODEN     #1{\unskip}     \fi
\ifx \showDOI      \undefined \def \showDOI       #1{#1}\fi
\ifx \showISBNx    \undefined \def \showISBNx     #1{\unskip}     \fi
\ifx \showISBNxiii \undefined \def \showISBNxiii  #1{\unskip}     \fi
\ifx \showISSN     \undefined \def \showISSN      #1{\unskip}     \fi
\ifx \showLCCN     \undefined \def \showLCCN      #1{\unskip}     \fi
\ifx \shownote     \undefined \def \shownote      #1{#1}          \fi
\ifx \showarticletitle \undefined \def \showarticletitle #1{#1}   \fi
\ifx \showURL      \undefined \def \showURL       {\relax}        \fi
\providecommand\bibfield[2]{#2}
\providecommand\bibinfo[2]{#2}
\providecommand\natexlab[1]{#1}
\providecommand\showeprint[2][]{arXiv:#2}

\bibitem[\protect\citeauthoryear{??}{sra}{2000}]%
        {sraa}
 \bibinfo{year}{1997-2000}\natexlab{}.
\newblock \bibinfo{title}{SRAA Dataset}.
\newblock
\newblock
\urldef\tempurl%
\url{https://people.cs.umass.edu/~mccallum/data.html}
\showURL{%
\tempurl}


\bibitem[\protect\citeauthoryear{??}{new}{2008}]%
        {newsgroups}
 \bibinfo{year}{2008}\natexlab{}.
\newblock \bibinfo{title}{Newsgroups Dataset}.
\newblock
\newblock
\urldef\tempurl%
\url{http://qwone.com/~jason/20Newsgroups/}
\showURL{%
\tempurl}


\bibitem[\protect\citeauthoryear{??}{imd}{2017}]%
        {imdb}
 \bibinfo{year}{2017}\natexlab{}.
\newblock \bibinfo{title}{IMDb Movie Dataset}.
\newblock
\newblock
\urldef\tempurl%
\url{ftp://ftp.fu-berlin.de/pub/misc/movies/database/}
\showURL{%
\tempurl}


\bibitem[\protect\citeauthoryear{??}{yah}{2018}]%
        {yahoo}
 \bibinfo{year}{2018}\natexlab{}.
\newblock \bibinfo{title}{Yahoo Movie Dataset}.
\newblock
\newblock
\urldef\tempurl%
\url{https://webscope.sandbox.yahoo.com/catalog.php?datatype=r}
\showURL{%
\tempurl}


\bibitem[\protect\citeauthoryear{Baldi, Cranmer, Faucett, Sadowski, and
  Whiteson}{Baldi et~al\mbox{.}}{2016}]%
        {baldi2016hepmass}
\bibfield{author}{\bibinfo{person}{Pierre Baldi}, \bibinfo{person}{Kyle
  Cranmer}, \bibinfo{person}{Taylor Faucett}, \bibinfo{person}{Peter Sadowski},
  {and} \bibinfo{person}{Daniel Whiteson}.} \bibinfo{year}{2016}\natexlab{}.
\newblock \showarticletitle{Parameterized neural networks for high-energy
  physics}.
\newblock \bibinfo{journal}{\emph{The European Physical Journal C}}
  \bibinfo{volume}{76} (\bibinfo{date}{05} \bibinfo{year}{2016}).
\newblock


\bibitem[\protect\citeauthoryear{Ben-David, Blitzer, Crammer, Kulesza, Pereira,
  and Vaughan}{Ben-David et~al\mbox{.}}{2010}]%
        {bendavid2010}
\bibfield{author}{\bibinfo{person}{Shai Ben-David}, \bibinfo{person}{John
  Blitzer}, \bibinfo{person}{Koby Crammer}, \bibinfo{person}{Alex Kulesza},
  \bibinfo{person}{Fernando Pereira}, {and} \bibinfo{person}{Jennifer~Wortman
  Vaughan}.} \bibinfo{year}{2010}\natexlab{}.
\newblock \showarticletitle{A theory of learning from different domains}.
\newblock \bibinfo{journal}{\emph{Machine Learning}} \bibinfo{volume}{79},
  \bibinfo{number}{1} (\bibinfo{date}{01 May} \bibinfo{year}{2010}),
  \bibinfo{pages}{151--175}.
\newblock


\bibitem[\protect\citeauthoryear{Bickel, Br\"{u}ckner, and Scheffer}{Bickel
  et~al\mbox{.}}{2009}]%
        {bickel2009covariate}
\bibfield{author}{\bibinfo{person}{Steffen Bickel}, \bibinfo{person}{Michael
  Br\"{u}ckner}, {and} \bibinfo{person}{Tobias Scheffer}.}
  \bibinfo{year}{2009}\natexlab{}.
\newblock \showarticletitle{Discriminative Learning Under Covariate Shift}.
\newblock \bibinfo{journal}{\emph{J. Mach. Learn. Res.}}  \bibinfo{volume}{10}
  (\bibinfo{date}{Dec.} \bibinfo{year}{2009}), \bibinfo{pages}{2137--2155}.
\newblock
\showISSN{1532-4435}


\bibitem[\protect\citeauthoryear{Blitzer, Dredze, and Pereira}{Blitzer
  et~al\mbox{.}}{2007}]%
        {blitzer2007amazon}
\bibfield{author}{\bibinfo{person}{John Blitzer}, \bibinfo{person}{Mark
  Dredze}, {and} \bibinfo{person}{Fernando Pereira}.}
  \bibinfo{year}{2007}\natexlab{}.
\newblock \showarticletitle{Biographies, Bollywood, Boom-boxes and Blenders:
  Domain Adaptation for Sentiment Classification}. In
  \bibinfo{booktitle}{\emph{Proceedings of the 45th Annual Meeting of the
  Association of Computational Linguistics}}. \bibinfo{publisher}{Association
  for Computational Linguistics}, \bibinfo{pages}{440--447}.
\newblock


\bibitem[\protect\citeauthoryear{Chapelle, Schlkopf, and Zien}{Chapelle
  et~al\mbox{.}}{2010}]%
        {chapelle2010semisupervised}
\bibfield{author}{\bibinfo{person}{Olivier Chapelle}, \bibinfo{person}{Bernhard
  Schlkopf}, {and} \bibinfo{person}{Alexander Zien}.}
  \bibinfo{year}{2010}\natexlab{}.
\newblock \bibinfo{booktitle}{\emph{Semi-Supervised Learning}
  (\bibinfo{edition}{1st} ed.)}.
\newblock \bibinfo{publisher}{The MIT Press}.
\newblock
\showISBNx{0262514125, 9780262514125}


\bibitem[\protect\citeauthoryear{Chen and Guestrin}{Chen and Guestrin}{2016}]%
        {chen2016xgboost}
\bibfield{author}{\bibinfo{person}{Tianqi Chen} {and} \bibinfo{person}{Carlos
  Guestrin}.} \bibinfo{year}{2016}\natexlab{}.
\newblock \showarticletitle{XGBoost: A Scalable Tree Boosting System}. In
  \bibinfo{booktitle}{\emph{KDD}}. \bibinfo{publisher}{ACM},
  \bibinfo{address}{New York, NY, USA}, \bibinfo{pages}{785--794}.
\newblock


\bibitem[\protect\citeauthoryear{Chu, Ho, Tsai, Lin, and Lin}{Chu
  et~al\mbox{.}}{2015}]%
        {chu2015warm}
\bibfield{author}{\bibinfo{person}{Bo-Yu Chu}, \bibinfo{person}{Chia-Hua Ho},
  \bibinfo{person}{Cheng-Hao Tsai}, \bibinfo{person}{Chieh-Yen Lin}, {and}
  \bibinfo{person}{Chih-Jen Lin}.} \bibinfo{year}{2015}\natexlab{}.
\newblock \showarticletitle{Warm start for parameter selection of linear
  classifiers}. In \bibinfo{booktitle}{\emph{KDD}}. ACM,
  \bibinfo{pages}{149--158}.
\newblock


\bibitem[\protect\citeauthoryear{Dai, Yang, Xue, and Yu}{Dai
  et~al\mbox{.}}{2007}]%
        {dai2007boosting}
\bibfield{author}{\bibinfo{person}{Wenyuan Dai}, \bibinfo{person}{Qiang Yang},
  \bibinfo{person}{Gui-Rong Xue}, {and} \bibinfo{person}{Yong Yu}.}
  \bibinfo{year}{2007}\natexlab{}.
\newblock \showarticletitle{Boosting for Transfer Learning}. In
  \bibinfo{booktitle}{\emph{ICML}}. \bibinfo{pages}{193--200}.
\newblock


\bibitem[\protect\citeauthoryear{Daum{\'e}}{Daum{\'e}}{2007}]%
        {daume2007easyda}
\bibfield{author}{\bibinfo{person}{Hal Daum{\'e}, III}.}
  \bibinfo{year}{2007}\natexlab{}.
\newblock \showarticletitle{Frustratingly Easy Domain Adaptation}. In
  \bibinfo{booktitle}{\emph{Proceedings of the 45th Annual Meeting of the
  Association of Computational Linguistics}}. \bibinfo{publisher}{Association
  for Computational Linguistics}, \bibinfo{pages}{256--263}.
\newblock


\bibitem[\protect\citeauthoryear{Daum{\'e} and Marcu}{Daum{\'e} and
  Marcu}{2006}]%
        {daume2006domainstat}
\bibfield{author}{\bibinfo{person}{Hal Daum{\'e}, III} {and}
  \bibinfo{person}{Daniel Marcu}.} \bibinfo{year}{2006}\natexlab{}.
\newblock \showarticletitle{Domain Adaptation for Statistical Classifiers}.
\newblock \bibinfo{journal}{\emph{J. Artif. Int. Res.}} \bibinfo{volume}{26},
  \bibinfo{number}{1} (\bibinfo{date}{May} \bibinfo{year}{2006}),
  \bibinfo{pages}{101--126}.
\newblock


\bibitem[\protect\citeauthoryear{{Deng}, {Dong}, {Socher}, {Li}, and
  and}{{Deng} et~al\mbox{.}}{2009}]%
        {deng2009imagenet}
\bibfield{author}{\bibinfo{person}{J. {Deng}}, \bibinfo{person}{W. {Dong}},
  \bibinfo{person}{R. {Socher}}, \bibinfo{person}{L. {Li}}, {and}
  \bibinfo{person}{and}.} \bibinfo{year}{2009}\natexlab{}.
\newblock \showarticletitle{ImageNet: A large-scale hierarchical image
  database}. In \bibinfo{booktitle}{\emph{CVPR}}. \bibinfo{pages}{248--255}.
\newblock
\showISSN{1063-6919}


\bibitem[\protect\citeauthoryear{Dua and Karra~Taniskidou}{Dua and
  Karra~Taniskidou}{2017}]%
        {dua2017uci}
\bibfield{author}{\bibinfo{person}{Dheeru Dua} {and} \bibinfo{person}{Efi
  Karra~Taniskidou}.} \bibinfo{year}{2017}\natexlab{}.
\newblock \bibinfo{title}{{UCI} Machine Learning Repository}.
\newblock
\newblock
\urldef\tempurl%
\url{http://archive.ics.uci.edu/ml}
\showURL{%
\tempurl}


\bibitem[\protect\citeauthoryear{{Halevy}, {Norvig}, and {Pereira}}{{Halevy}
  et~al\mbox{.}}{2009}]%
        {halevy2009data}
\bibfield{author}{\bibinfo{person}{A. {Halevy}}, \bibinfo{person}{P. {Norvig}},
  {and} \bibinfo{person}{F. {Pereira}}.} \bibinfo{year}{2009}\natexlab{}.
\newblock \showarticletitle{The Unreasonable Effectiveness of Data}.
\newblock \bibinfo{journal}{\emph{IEEE Intelligent Systems}}
  \bibinfo{volume}{24}, \bibinfo{number}{2} (\bibinfo{date}{March}
  \bibinfo{year}{2009}), \bibinfo{pages}{8--12}.
\newblock


\bibitem[\protect\citeauthoryear{Huang, Li, Yu, Deng, and Gong}{Huang
  et~al\mbox{.}}{2013}]%
        {tlsharednn}
\bibfield{author}{\bibinfo{person}{Jui-Ting Huang}, \bibinfo{person}{Jinyu Li},
  \bibinfo{person}{Dong Yu}, \bibinfo{person}{Li Deng}, {and}
  \bibinfo{person}{Yifan Gong}.} \bibinfo{year}{2013}\natexlab{}.
\newblock \showarticletitle{Cross-language knowledge transfer using
  multilingual deep neural network with shared hidden layers}. In
  \bibinfo{booktitle}{\emph{2013 IEEE International Conference on Acoustics,
  Speech and Signal Processing}}. IEEE, \bibinfo{pages}{7304--7308}.
\newblock


\bibitem[\protect\citeauthoryear{Kiefer}{Kiefer}{1953}]%
        {gss}
\bibfield{author}{\bibinfo{person}{Jack Kiefer}.}
  \bibinfo{year}{1953}\natexlab{}.
\newblock \showarticletitle{Sequential minimax search for a maximum}.
\newblock \bibinfo{journal}{\emph{Proceedings of the American mathematical
  society}} \bibinfo{volume}{4}, \bibinfo{number}{3} (\bibinfo{year}{1953}),
  \bibinfo{pages}{502--506}.
\newblock


\bibitem[\protect\citeauthoryear{Klein, Falkner, Bartels, Hennig, and
  Hutter}{Klein et~al\mbox{.}}{2016}]%
        {klein2016fast}
\bibfield{author}{\bibinfo{person}{Aaron Klein}, \bibinfo{person}{Stefan
  Falkner}, \bibinfo{person}{Simon Bartels}, \bibinfo{person}{Philipp Hennig},
  {and} \bibinfo{person}{Frank Hutter}.} \bibinfo{year}{2016}\natexlab{}.
\newblock \showarticletitle{Fast bayesian optimization of machine learning
  hyperparameters on large datasets}.
\newblock \bibinfo{journal}{\emph{arXiv preprint arXiv:1605.07079}}
  (\bibinfo{year}{2016}).
\newblock


\bibitem[\protect\citeauthoryear{Krishnan, Wang, Wu, Franklin, and
  Goldberg}{Krishnan et~al\mbox{.}}{2016}]%
        {krishnan2016activeclean}
\bibfield{author}{\bibinfo{person}{Sanjay Krishnan}, \bibinfo{person}{Jiannan
  Wang}, \bibinfo{person}{Eugene Wu}, \bibinfo{person}{Michael~J Franklin},
  {and} \bibinfo{person}{Ken Goldberg}.} \bibinfo{year}{2016}\natexlab{}.
\newblock \showarticletitle{Activeclean: Interactive data cleaning for
  statistical modeling}.
\newblock \bibinfo{journal}{\emph{Proceedings of the VLDB Endowment}}
  \bibinfo{volume}{9}, \bibinfo{number}{12} (\bibinfo{year}{2016}),
  \bibinfo{pages}{948--959}.
\newblock


\bibitem[\protect\citeauthoryear{{Lecun}, {Bottou}, {Bengio}, and
  {Haffner}}{{Lecun} et~al\mbox{.}}{1998}]%
        {lecun1998mnist}
\bibfield{author}{\bibinfo{person}{Y. {Lecun}}, \bibinfo{person}{L. {Bottou}},
  \bibinfo{person}{Y. {Bengio}}, {and} \bibinfo{person}{P. {Haffner}}.}
  \bibinfo{year}{1998}\natexlab{}.
\newblock \showarticletitle{Gradient-based learning applied to document
  recognition}.
\newblock \bibinfo{journal}{\emph{Proc. IEEE}} \bibinfo{volume}{86},
  \bibinfo{number}{11} (\bibinfo{date}{Nov} \bibinfo{year}{1998}),
  \bibinfo{pages}{2278--2324}.
\newblock


\bibitem[\protect\citeauthoryear{Li, Jamieson, DeSalvo, Rostamizadeh, and
  Talwalkar}{Li et~al\mbox{.}}{2017}]%
        {li2016hyperband}
\bibfield{author}{\bibinfo{person}{Lisha Li}, \bibinfo{person}{Kevin Jamieson},
  \bibinfo{person}{Giulia DeSalvo}, \bibinfo{person}{Afshin Rostamizadeh},
  {and} \bibinfo{person}{Ameet Talwalkar}.} \bibinfo{year}{2017}\natexlab{}.
\newblock \showarticletitle{Hyperband: A Novel Bandit-based Approach to
  Hyperparameter Optimization}.
\newblock \bibinfo{journal}{\emph{J. Mach. Learn. Res.}} \bibinfo{volume}{18},
  \bibinfo{number}{1} (\bibinfo{date}{Jan.} \bibinfo{year}{2017}),
  \bibinfo{pages}{6765--6816}.
\newblock


\bibitem[\protect\citeauthoryear{Loosli, Canu, and Bottou}{Loosli
  et~al\mbox{.}}{2007}]%
        {loosli2006mnistinifinite}
\bibfield{author}{\bibinfo{person}{Ga\"{e}lle Loosli},
  \bibinfo{person}{St\'{e}phane Canu}, {and} \bibinfo{person}{L\'{e}on
  Bottou}.} \bibinfo{year}{2007}\natexlab{}.
\newblock \showarticletitle{Training Invariant Support Vector Machines using
  Selective Sampling}.
\newblock In \bibinfo{booktitle}{\emph{Large Scale Kernel Machines}},
  \bibfield{editor}{\bibinfo{person}{L\'{e}on Bottou}, \bibinfo{person}{Olivier
  Chapelle}, \bibinfo{person}{Dennis {DeCoste}}, {and} \bibinfo{person}{Jason
  Weston}} (Eds.). \bibinfo{publisher}{MIT Press}, \bibinfo{address}{Cambridge,
  MA.}, \bibinfo{pages}{301--320}.
\newblock
\urldef\tempurl%
\url{http://leon.bottou.org/papers/loosli-canu-bottou-2006}
\showURL{%
\tempurl}


\bibitem[\protect\citeauthoryear{McAuley and Leskovec}{McAuley and
  Leskovec}{2013}]%
        {mcauley2013ratingreview}
\bibfield{author}{\bibinfo{person}{Julian McAuley} {and} \bibinfo{person}{Jure
  Leskovec}.} \bibinfo{year}{2013}\natexlab{}.
\newblock \showarticletitle{Hidden Factors and Hidden Topics: Understanding
  Rating Dimensions with Review Text}. In \bibinfo{booktitle}{\emph{ACM
  Conference on Recommender Systems}} \emph{(\bibinfo{series}{RecSys '13})}.
  \bibinfo{pages}{165--172}.
\newblock


\bibitem[\protect\citeauthoryear{Pan and Yang}{Pan and Yang}{2010}]%
        {pan2010transfersurvey}
\bibfield{author}{\bibinfo{person}{Sinno~Jialin Pan} {and}
  \bibinfo{person}{Qiang Yang}.} \bibinfo{year}{2010}\natexlab{}.
\newblock \showarticletitle{A survey on transfer learning}.
\newblock \bibinfo{journal}{\emph{IEEE Transactions on knowledge and data
  engineering}} \bibinfo{volume}{22}, \bibinfo{number}{10}
  (\bibinfo{year}{2010}), \bibinfo{pages}{1345--1359}.
\newblock


\bibitem[\protect\citeauthoryear{Pedregosa, Varoquaux, Gramfort, Michel,
  Thirion, Grisel, Blondel, Prettenhofer, Weiss, Dubourg, Vanderplas, Passos,
  Cournapeau, Brucher, Perrot, and Duchesnay}{Pedregosa et~al\mbox{.}}{2011}]%
        {scikit-learn}
\bibfield{author}{\bibinfo{person}{F. Pedregosa}, \bibinfo{person}{G.
  Varoquaux}, \bibinfo{person}{A. Gramfort}, \bibinfo{person}{V. Michel},
  \bibinfo{person}{B. Thirion}, \bibinfo{person}{O. Grisel},
  \bibinfo{person}{M. Blondel}, \bibinfo{person}{P. Prettenhofer},
  \bibinfo{person}{R. Weiss}, \bibinfo{person}{V. Dubourg}, \bibinfo{person}{J.
  Vanderplas}, \bibinfo{person}{A. Passos}, \bibinfo{person}{D. Cournapeau},
  \bibinfo{person}{M. Brucher}, \bibinfo{person}{M. Perrot}, {and}
  \bibinfo{person}{E. Duchesnay}.} \bibinfo{year}{2011}\natexlab{}.
\newblock \showarticletitle{Scikit-learn: Machine Learning in {P}ython}.
\newblock \bibinfo{journal}{\emph{Journal of Machine Learning Research}}
  \bibinfo{volume}{12} (\bibinfo{year}{2011}), \bibinfo{pages}{2825--2830}.
\newblock


\bibitem[\protect\citeauthoryear{Ratner, Bach, Ehrenberg, Fries, Wu, and
  R{\'e}}{Ratner et~al\mbox{.}}{2017}]%
        {snorkel}
\bibfield{author}{\bibinfo{person}{Alexander Ratner},
  \bibinfo{person}{Stephen~H Bach}, \bibinfo{person}{Henry Ehrenberg},
  \bibinfo{person}{Jason Fries}, \bibinfo{person}{Sen Wu}, {and}
  \bibinfo{person}{Christopher R{\'e}}.} \bibinfo{year}{2017}\natexlab{}.
\newblock \showarticletitle{Snorkel: Rapid training data creation with weak
  supervision}.
\newblock \bibinfo{journal}{\emph{Proceedings of the VLDB Endowment}}
  \bibinfo{volume}{11}, \bibinfo{number}{3} (\bibinfo{year}{2017}),
  \bibinfo{pages}{269--282}.
\newblock


\bibitem[\protect\citeauthoryear{Rekatsinas, Chu, Ilyas, and R{\'e}}{Rekatsinas
  et~al\mbox{.}}{2017}]%
        {holoclean}
\bibfield{author}{\bibinfo{person}{Theodoros Rekatsinas}, \bibinfo{person}{Xu
  Chu}, \bibinfo{person}{Ihab~F Ilyas}, {and} \bibinfo{person}{Christopher
  R{\'e}}.} \bibinfo{year}{2017}\natexlab{}.
\newblock \showarticletitle{Holoclean: Holistic data repairs with probabilistic
  inference}.
\newblock \bibinfo{journal}{\emph{Proceedings of the VLDB Endowment}}
  \bibinfo{volume}{10}, \bibinfo{number}{11} (\bibinfo{year}{2017}),
  \bibinfo{pages}{1190--1201}.
\newblock


\bibitem[\protect\citeauthoryear{Robbins and Monro}{Robbins and Monro}{1951}]%
        {robbins1951sgd}
\bibfield{author}{\bibinfo{person}{Herbert Robbins} {and}
  \bibinfo{person}{Sutton Monro}.} \bibinfo{year}{1951}\natexlab{}.
\newblock \showarticletitle{A Stochastic Approximation Method}.
\newblock \bibinfo{journal}{\emph{The Annals of Mathematical Statistics}}
  \bibinfo{volume}{22}, \bibinfo{number}{3} (\bibinfo{year}{1951}),
  \bibinfo{pages}{400--407}.
\newblock


\bibitem[\protect\citeauthoryear{Settles}{Settles}{2012}]%
        {burr2012active}
\bibfield{author}{\bibinfo{person}{Burr Settles}.}
  \bibinfo{year}{2012}\natexlab{}.
\newblock \showarticletitle{Active Learning}.
\newblock \bibinfo{journal}{\emph{Synthesis Lectures on Artificial Intelligence
  and Machine Learning}} \bibinfo{volume}{6}, \bibinfo{number}{1}
  (\bibinfo{year}{2012}), \bibinfo{pages}{1--114}.
\newblock


\bibitem[\protect\citeauthoryear{Shimodaira}{Shimodaira}{2000}]%
        {shimodaira2000covariateshift}
\bibfield{author}{\bibinfo{person}{Hidetoshi Shimodaira}.}
  \bibinfo{year}{2000}\natexlab{}.
\newblock \showarticletitle{Improving predictive inference under covariate
  shift by weighting the log-likelihood function}.
\newblock \bibinfo{journal}{\emph{Journal of Statistical Planning and
  Inference}} \bibinfo{volume}{90}, \bibinfo{number}{2} (\bibinfo{year}{2000}),
  \bibinfo{pages}{227 -- 244}.
\newblock


\bibitem[\protect\citeauthoryear{Sun, Feng, and Saenko}{Sun
  et~al\mbox{.}}{2016}]%
        {sun2016coral}
\bibfield{author}{\bibinfo{person}{Baochen Sun}, \bibinfo{person}{Jiashi Feng},
  {and} \bibinfo{person}{Kate Saenko}.} \bibinfo{year}{2016}\natexlab{}.
\newblock \showarticletitle{Return of Frustratingly Easy Domain Adaptation}. In
  \bibinfo{booktitle}{\emph{AAAI}} \emph{(\bibinfo{series}{AAAI'16})}.
  \bibinfo{pages}{2058--2065}.
\newblock


\bibitem[\protect\citeauthoryear{Sun, Shrivastava, Singh, and Gupta}{Sun
  et~al\mbox{.}}{2017}]%
        {sun2017revisitdata}
\bibfield{author}{\bibinfo{person}{Chen Sun}, \bibinfo{person}{Abhinav
  Shrivastava}, \bibinfo{person}{Saurabh Singh}, {and} \bibinfo{person}{Abhinav
  Gupta}.} \bibinfo{year}{2017}\natexlab{}.
\newblock \showarticletitle{Revisiting Unreasonable Effectiveness of Data in
  Deep Learning Era.}. In \bibinfo{booktitle}{\emph{ICCV}}.
  \bibinfo{pages}{843--852}.
\newblock


\bibitem[\protect\citeauthoryear{Tan, Sun, Kong, Zhang, Yang, and Liu}{Tan
  et~al\mbox{.}}{2018}]%
        {deeptransfersurvey}
\bibfield{author}{\bibinfo{person}{Chuanqi Tan}, \bibinfo{person}{Fuchun Sun},
  \bibinfo{person}{Tao Kong}, \bibinfo{person}{Wenchang Zhang},
  \bibinfo{person}{Chao Yang}, {and} \bibinfo{person}{Chunfang Liu}.}
  \bibinfo{year}{2018}\natexlab{}.
\newblock \showarticletitle{A Survey on Deep Transfer Learning}.
\newblock \bibinfo{journal}{\emph{CoRR}}  \bibinfo{volume}{abs/1808.01974}
  (\bibinfo{year}{2018}).
\newblock
\showeprint{1808.01974}


\bibitem[\protect\citeauthoryear{Tommasi, Orabona, and Caputo}{Tommasi
  et~al\mbox{.}}{2010}]%
        {tommasi2010safety}
\bibfield{author}{\bibinfo{person}{Tatiana Tommasi}, \bibinfo{person}{Francesco
  Orabona}, {and} \bibinfo{person}{Barbara Caputo}.}
  \bibinfo{year}{2010}\natexlab{}.
\newblock \showarticletitle{Safety in numbers: Learning categories from few
  examples with multi model knowledge transfer}. In
  \bibinfo{booktitle}{\emph{CVPR}}. \bibinfo{pages}{3081--3088}.
\newblock


\bibitem[\protect\citeauthoryear{Tsai, Lin, and Lin}{Tsai
  et~al\mbox{.}}{2014}]%
        {tsai2014incremental}
\bibfield{author}{\bibinfo{person}{Cheng-Hao Tsai}, \bibinfo{person}{Chieh-Yen
  Lin}, {and} \bibinfo{person}{Chih-Jen Lin}.} \bibinfo{year}{2014}\natexlab{}.
\newblock \showarticletitle{Incremental and decremental training for linear
  classification}. In \bibinfo{booktitle}{\emph{KDD}}.
  \bibinfo{pages}{343--352}.
\newblock


\bibitem[\protect\citeauthoryear{Vergara, Huerta, Ayhan, Ryan, Vembu, and
  Homer}{Vergara et~al\mbox{.}}{2011}]%
        {vergara2011gas}
\bibfield{author}{\bibinfo{person}{Alexander Vergara},
  \bibinfo{person}{Ram\'{o}n Huerta}, \bibinfo{person}{Tuba Ayhan},
  \bibinfo{person}{Margaret Ryan}, \bibinfo{person}{Shankar Vembu}, {and}
  \bibinfo{person}{Margie Homer}.} \bibinfo{year}{2011}\natexlab{}.
\newblock \showarticletitle{Gas Sensor Drift Mitigation Using Classifier
  Ensembles}. In \bibinfo{booktitle}{\emph{SensorKDD}}.
  \bibinfo{pages}{16--24}.
\newblock


\bibitem[\protect\citeauthoryear{Weiss, Khoshgoftaar, and Wang}{Weiss
  et~al\mbox{.}}{2016}]%
        {weiss2016survey}
\bibfield{author}{\bibinfo{person}{Karl Weiss}, \bibinfo{person}{Taghi~M
  Khoshgoftaar}, {and} \bibinfo{person}{DingDing Wang}.}
  \bibinfo{year}{2016}\natexlab{}.
\newblock \showarticletitle{A survey of transfer learning}.
\newblock \bibinfo{journal}{\emph{Journal of Big Data}} \bibinfo{volume}{3},
  \bibinfo{number}{1} (\bibinfo{year}{2016}), \bibinfo{pages}{9}.
\newblock


\bibitem[\protect\citeauthoryear{Yosinski, Clune, Bengio, and Lipson}{Yosinski
  et~al\mbox{.}}{2014}]%
        {yosinski2014transferable}
\bibfield{author}{\bibinfo{person}{Jason Yosinski}, \bibinfo{person}{Jeff
  Clune}, \bibinfo{person}{Yoshua Bengio}, {and} \bibinfo{person}{Hod Lipson}.}
  \bibinfo{year}{2014}\natexlab{}.
\newblock \showarticletitle{How transferable are features in deep neural
  networks?}. In \bibinfo{booktitle}{\emph{NIPS}}. \bibinfo{pages}{3320--3328}.
\newblock


\end{thebibliography}

\end{document}